\documentclass[letterpaper, 10 pt, conference]{ieeeconf}  

\IEEEoverridecommandlockouts                              

\usepackage{epsfig} 
\usepackage{amsmath} 
\usepackage{amssymb}  

\usepackage{bm}
\usepackage{caption}
\usepackage{subcaption}
\usepackage{hyperref}

\title{{\scriptsize Copyright is transferred to IEEE, DOI: \href{https://ieeexplore.ieee.org/document/9867812}{10.23919/ACC53348.2022.9867812}}
	
	\LARGE \bf
Robot Control for Simultaneous Impact tasks via \\ Quadratic Programming-based Reference Spreading
}

\author{Jari J. van Steen, Nathan van de Wouw and Alessandro Saccon
}

\begin{document}

\maketitle
\thispagestyle{empty}
\pagestyle{empty}

\begin{abstract}

With the aim of further enabling the exploitation of impacts in robotic manipulation, a control framework is presented that directly tackles the challenges posed by tracking control of robotic manipulators that are tasked to perform nominally simultaneous impacts associated to multiple contact points. To this end, we extend the framework of reference spreading, which uses an extended ante- and post-impact reference coherent with a rigid impact map, determined under the assumption of an inelastic simultaneous impact. In practice, the robot will not reside exactly on the reference at the impact moment; as a result a sequence of impacts at the different contact points will typically occur. Our new approach extends reference spreading in this context via the introduction of an additional interim control mode. In this mode, a torque command is still based on the ante-impact reference with the goal of reaching the target contact state, but velocity feedback is disabled as this can be potentially harmful due to rapid velocity changes. With an eye towards real implementation, the approach is formulated using a quadratic programming (QP) control framework and is validated using numerical simulations both on a rigid robot model and on a realistic robot model with flexible joints.

\end{abstract}

\section{Introduction}\label{sec:introduction} 

	Humans are extremely skilled in using collisions to perform or speed up the execution of motion tasks. This can be seen in activities like running or kicking a ball, and even in more basic tasks like opening a door or grabbing an object. While straightforward for most humans, this skill is challenging to translate to robots. State-of-the-art robot control \cite{Salehian2016} enforces almost-zero relative speed between the end effector and the object or environment upon contact. While this approach is viable and sometimes a necessity, there are various applications where exploiting impacts leads to faster motions, potentially also being more robust.
	The exploitation of impacts has been and is still an active area of research in robot locomotion \cite{Gu2018,Hutter2011,Westervelt2002}, while it is beginning to be explored also in the context of robot manipulation with objects of non-negligible weight \cite{Stouraitis2020}. 
	 
	In this work, we explore the problem of tracking control in the presence of nominally simultaneous impacts for a robotic manipulator, considering tracking control specifically as an approach to enable the execution of tasks with high precision, where the impact time is also of relevance. 
	Tracking control of a reference with impact-induced jumps poses a challenge around the time of impact, as the velocity tracking error will peak around the nominal impact time as a result of a mismatch between the actual impact time and the nominal impact under regular tracking control \cite{Biemond2013,Forni2013,Leine2008}. 	
	An additional challenge arises when several impacts are planned to be executed simultaneously \cite{Pace2017} as we do in this paper. An example where such a motion is desired is dual-arm manipulation \cite{Amanhoud2019}, which can be applied in logistic applications, such as automated palletizing or depalletizing, where being able to utilize nominally simultaneous impacts instead of making contact at almost zero velocity could significantly speed up such processes. 
	However, inevitable misalignment of the impacting surface(s) of the robot and the environment leads to spurious impacts to take place, resulting in uncertain contact states with imprecise velocity measurements, which complicates tracking control when compared with a single impact scenario. 
	The goal of this paper is hence to enable tracking control in mechanical systems that experience nominally simultaneous impacts.

	Previous work on tracking control of references with single and/or simultaneous impacts includes \cite{Sanfelice2014}, which describes a hybrid control approach where it is assumed that the impact in the reference and in the real system occur simultaneously. 
	The approach taken in \cite{Biemond2013} works around the error peaking phenomenon by defining a distance function based on the predicted state jump to formulate the tracking error in such a way that the mismatch between the predicted and the actual jump times does not have an effect on the tracking error. 
	Alternatively, \cite{Yang2021} actively avoids peaks in the velocity error by projecting the tracking tasks onto a subspace that is invariant to the impact event around the impact time, meaning that the velocity jump will not be observed. 
	In \cite{Morarescu2010}, a so-called transition phase is introduced to apply tracking control during contact transitions (impacts and detachment).
	
	Another approach that has been proposed to perform tracking control with impacts, which we will extend in this work, is the reference spreading framework, introduced in \cite{Saccon2014} and further developed in \cite{Rijnen2015,Rijnen2017,Rijnen2019a}. This framework enables tracking control of mechanical systems through a hybrid control approach with an ante- and post-impact reference that overlap about the nominal impact time. The switching from ante- to post-impact control mode is done based on detection of the impact, instead of being tied to the nominal impact time, to remove the peak in the velocity error, which can lead to unpredictable behavior. Experimental validation of reference spreading on a physical setup has been reported in \cite{Rijnen2020}. 
	When dealing with robotic and mechanical systems, reference spreading is applied by modeling contacts as rigid and impacts as inelastic, following the theory of multibody dynamics and nonsmooth mechanics \cite{Brogliato2016,Glocker2006} to capture the configuration dependent velocity jumps by means of impact laws \cite{Rijnen2017}. 
	Although designed for rigid contacts and rigid robots, the reference spreading framework actually works and is intended to be applied to robots, environments, and objects that exhibit flexibility (which is unavoidable, from a practical perspective), as is also demonstrated in this paper by means of numerical simulations.  
	
	Reference spreading is currently formulated as a full-state feedback control approach. It is however more common in robotic applications to specify control tasks in Cartesian space. To show that taking a Cartesian perspective is also possible, we here cast reference spreading in a quadratic programming (QP) control framework \cite{Bouyarmane2019,Salini2010}, where the control input is obtained from a linearly constrained quadratic optimization problem. QP control can be used for control of robots with multiple tasks in joint space or operational space, under a set of constraints ensuring, for example, adherence to joint limits or avoidance of unwanted collisions. 
	Development of impact-aware QP control is an active area of research: \cite{Wang2019}, e.g., takes the perspective of robot safety, focusing on slowing down the robot to ensure that an expected impact will stay within hardware capabilities. 
	
	Reference spreading control for nominally simultaneous impacts has been addressed in \cite{Rijnen2019}, focusing on a sensitivity analysis. The approach was numerically validated on a controlled box impacting and sliding on a compliant hinged plank. In \cite{Rijnen2019}, an intermediate control mode was also proposed in the time frame when contact between the box and the plank has been partially established, making use of pure feedforward during the intermediate mode. In this work, instead, we show the additional use of position feedback and we verify the approach in simulation. We plan the reference trajectory employing a fully rigid robot and environment while demonstrating the success of the control strategy also on a realistic flexible joint robot with compliant contact model. Furthermore, the approach is integrated within the QP robot control framework under the assumption that there is no task redundancy in the ante-impact mode. 

	The paper is structured as follows. In section \ref{sec:robot_dynamics}, the dynamics of the manipulator that will be used to demonstrate the control approach are presented. In section \ref{sec:reference_generation}, the extended ante- and post-impact reference will be described, using the framework of reference spreading. In section \ref{sec:control_approach}, the control approach itself is formulated, followed by a numerical validation on a rigid and flexible robot model in Section \ref{sec:numerical_validation}. Finally, the conclusions are presented in Section \ref{sec:conclusion}.

\section{Robot dynamics}\label{sec:robot_dynamics}

\begin{figure}
		\centering
		\includegraphics[width=0.9\linewidth]{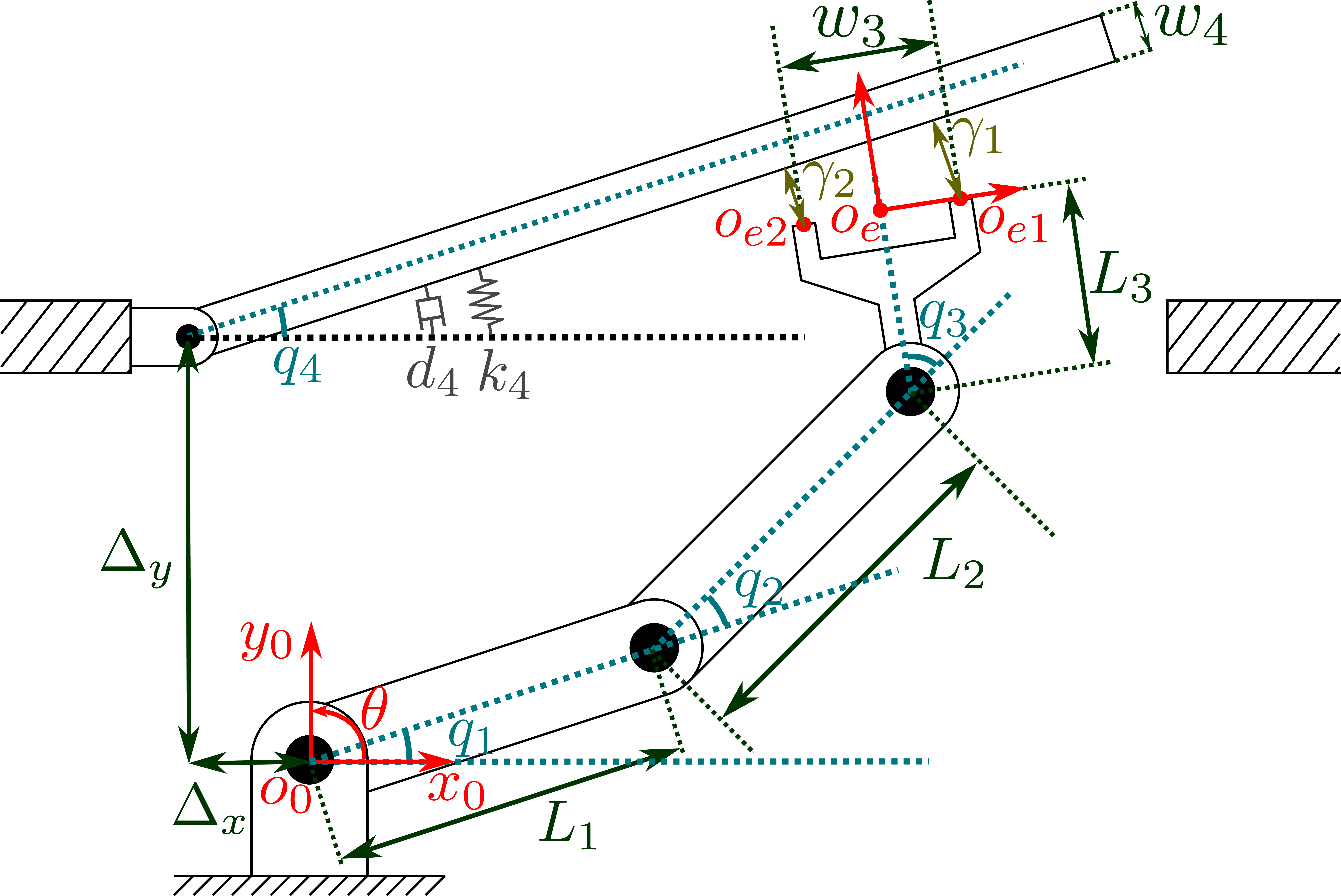}
		\caption{Overview of the 3DOF planar manipulator impacting a hinged rigid plank}\label{fig:robot_3DOF}
	\end{figure}

	While the control strategy described in this paper can be applied to a wider range of scenarios, the planar manipulator with three degrees of freedom (DOF) depicted in Figure \ref{fig:robot_3DOF} is used throughout this paper to illustrate and demonstrate the approach. The example presents all key aspects of the problem, allowing us to be concrete without the need of becoming too onerous in terms of notation. 
	The robot consists of three rigid frictionless actuated joints and rigid links, and impacts a hinged rigid plank. The generalized coordinate vector of the system is $\bm q = \left[\bm q_\text{rob}^T \ q_4\right]^T$ with $\bm q_\text{rob} = \left[q_1 \ q_2 \ q_3\right]^T$. The mass of each link $i$ and its inertia around the center of gravity are given by $m_i$ and $I_{g,i}$ respectively, with the inertia of the plank around the hinge given by $I_{o,4}$. 
	The end effector position $\bm p = \left[x_e \ y_e\right]^T$ and orientation $\theta$ describe the position and orientation of frame $e$ expressed in terms of frame $0$, with their respective velocities given by 
	\begin{equation}
		\dot{\bm p} = \bm J_p(\bm q) \dot{\bm{q}}, \ \ \ \dot{\theta} = \bm J_\theta(\bm q) \dot{\bm{q}}.
	\end{equation}
	with $\bm J_p(\bm q) = \left[\bm J_{p,\text{rob}}(\bm q) \ \bm 0\right]$, $\bm J_\theta(\bm q) = \left[\bm J_{\theta,\text{rob}}(\bm q) \  0\right]$, and
	$$
	\bm J_{p,\text{rob}}(\bm q) = \frac{\partial \bm p}{\partial \bm q_\text{rob}},	\ \ \	\bm J_{\theta,\text{rob}}(\bm q) = \frac{\partial \theta}{\partial \bm q_\text{rob}}.
	$$
	For ease of notation, we will drop the explicit dependency on $\bm q$ (or $\dot{\bm q}$). The contact between the end effector and the plank is assumed to be frictionless, and can be described by the gap functions $\gamma_i$ for $i \in\{1,2\}$ and corresponding contact force $\lambda_i$ adhering to the complementarity condition 
	\begin{equation}\label{eq:complementarity}
		0 \leq \gamma_i \perp \lambda_i \geq 0,
	\end{equation}
	which is a common description in the framework of nonsmooth mechanics \cite{Brogliato2016}. This implies that the contact force is zero when the contact is open and non-negative when the contact is closed.
	\subsection{Non-impulsive free and constrained dynamics}
	During free or constrained motion, the robot dynamics are described by
	\begin{equation}\label{eq:EOM}
	\bm M \ddot{\bm q} + \bm h = \bm S \bm \tau + \bm J_{N}^T  \bm \lambda,
	\end{equation}
	with joint positions $\bm q$, mass matrix $\bm M$, vector of gravity, centrifugal and Coriolis terms $\bm h$, applied joint torques $\bm \tau$, actuation matrix $\bm S = \left[\bm I_{3 \times 3} \ \bm 0_3 \right]^T$, and normal contact forces $\bm \lambda$ (being zero in free motion). The contact Jacobian is given by $\bm J_N = \left[\bm J_{N,1}^T \ \bm J_{N,2}^T\right]^T$, with
	\begin{equation}
	\bm J_{N,i} = \frac{\partial \gamma_i}{\partial \bm q}.
	\end{equation}

	\subsection{Impulsive impact dynamics}
	
	To ensure adherence to the non-compenetration condition $\gamma_i \geq0$ appearing in \eqref{eq:complementarity}, an instantaneous jump in $\dot{\bm q}$ has to be allowed as soon as a contact is closed, which is described by the so-called impact map \cite{Glocker2006}. This results in a potential discontinuity in joint velocities at the impact time, with ante-impact velocity $\dot{\bm q}^-$, post-impact velocity $\dot{\bm q}^+$, and continuous position $\bm q^- = \bm q^+$. Integrating \eqref{eq:EOM} over the impact time leads to the impact equation \cite{Brogliato2016} given by
	\begin{equation}\label{eq:impact_eq_new}
	\bm M\left(\dot{\bm q}^+ - \dot{\bm q}^{-}\right) = \bm J_{N}^T \bm \Lambda,
	\end{equation}
	with $\bf{\bm \Lambda}$ the momentum associated to the impulsive contact forces. As mentioned in the introduction, we assume  inelastic impacts to take place at the moment of contact transition ($\dot{\gamma}_i^+ = 0$). This means that, when contact of the end effector with the plank is established simultaneously at both possible contact locations, we have 
	\begin{equation}\label{eq:impact_law_new}
	\bm J_{N} \dot{\bm q}^+ = \bm 0.
	\end{equation}	
	Combining \eqref{eq:impact_eq_new} and \eqref{eq:impact_law_new} leads to the simultaneous impact map 
	\begin{equation}\label{eq:impact_map} 
	\dot{\bm q}^+ = \left(\bm I - \bm M^{-1} \bm J_{N}^T \bm \left(\bm J_{N} \bm M^{-1} \bm J_{N}^T\right)^{-1}\bm J_{N}\right)\dot{\bm q}^{-},
	\end{equation}
	which will be used in the formulation of the reference in Section \ref{sec:reference_generation}.
	
	\section{Reference trajectory generation} \label{sec:reference_generation}
	
	The key aspect of reference spreading is the re-definition of the tracking error by means of employing two reference trajectories that are related via the impact map \eqref{eq:impact_map}, which partially overlap in time around the nominal impact moment, in order to have at a hand a suitable reference even when impacts occur at different times than planned.  
	In previous work \cite{Rijnen2017}, the ante and post-impact references are formulated in operational space, but are then converted and tracked in state space. In this work, we are instead providing and directly using the reference trajectory in operational space, under the assumption, mentioned in the introduction, that there is no task redundancy in the ante-impact reference.

	In the following, we illustrate the generation of the operational space reference trajectory for the planar manipulation scenario depicted in Figure \ref{fig:robot_3DOF}. The nominal motion of the manipulator is described by a desired ante-impact end effector position and orientation ${\bm p}^a_{d}(t)$ and ${\theta}^a_{d}(t)$, as well as a desired post-impact position ${\bm p}^p_{d}(t)$. Note that specifying a desired post-impact orientation of the end effector is not necessary, because, under the assumption of an inelastic impact and sustained contact with the plank, both the plank and end effector orientation are implicitly defined by the end effector position $\bm p_d^p (t)$. 
	To make the ante- and post-impact reference trajectories consistent, it must be ensured that at the nominal impact time $t_\text{imp}$, the nominal ante-impact state $(\bm q^-, \dot{\bm q}^-)$ relates to the nominal post-impact state $(\bm q^+, \dot{\bm q}^+)$ through the impact map \eqref{eq:impact_map}. In Appendix A, we provide the details on how to ensure this impact consistency for the scenario of Figure \ref{fig:robot_3DOF}.

	\begin{figure}		
		\centering
		\includegraphics[width=0.9\linewidth]{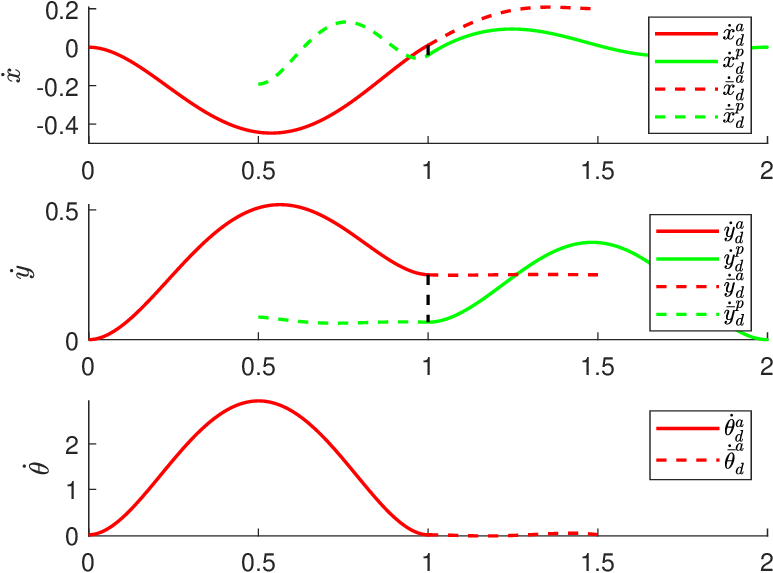}
		\caption{Depiction of the time derivative of the extended ante- and post-impact position reference $\dot{\bar{\bm p}}^a_{d}(t)$ and $\dot{\bar{\bm p}}^p_{d}(t)$, and ante-impact orientation reference $\dot{\bar{\theta}}^a_{d}(t)$ with nominal impact at time $t_\text{imp}=1$.}
		\label{fig:RS}
	\end{figure}
	
	Assuming to have at hand a reference trajectory which is consistent with the impact map \eqref{eq:impact_map}, we can proceed with the general recipe of reference spreading, namely, extension of these references by integrating the ante-impact reference forward and the post-impact reference backward \cite{Saccon2014}. This integration	results in the extended references $\bar{\bm p}^a_{d}(t)$, $\bar{\theta}^a_{d}(t)$, and $\bar{\bm p}^p_{d}(t)$, as visualized in Figure \ref{fig:RS} by means of dashed lines. Depending on the contact state of the manipulator, the corresponding reference will be used for feedback control, as will be explained in the Section \ref{sec:control_approach}.

\section{Control approach}\label{sec:control_approach}

	In Section \ref{sec:reference_generation}, a desired motion with nominally simultaneous impacts is obtained, illustrating the approach for the manipulator depicted in Figure \ref{fig:robot_3DOF}. In this section, we show how to construct a reference spreading based control action which is defined in operational space and makes use of QP robot control \cite{Bouyarmane2019,Salini2010}. In particular, we detail how the ante-impact, intermediate, and post-impact control actions can be realized via suitable QPs, obtaining for each mode a suitable joint torque $\bm \tau^*$ to be applied to the robot at every time step $\Delta t$. 
	The main objective in the ante- and post-impact mode is to track the corresponding reference defined in Section \ref{sec:reference_generation}, while the main objective of the intermediate mode is to ensure that full contact is established without explicit velocity feedback or knowledge of the contact state, both unreliable during the spurious sequence of impacts associated with the nominally simultaneous impact. The three QPs corresponding to the ante-impact, intermediate, and post-impact mode are detailed below.

\subsection{Ante-impact mode}

	In the ante-impact mode, as no contact force is present, the QP optimization variables are given by the torques $\bm \tau$ and the accelerations $\ddot{\bm q}$. 
	Assuming to measure $\bm q$ and obtain an estimate of $\dot{\bm q}$ at each time step, both are assumed to be known. 

	The cost function of the QP is described by means of a weighted sum of the costs corresponding to different tasks that are to be executed. By minimizing this cost function under the yet-to-be-defined constraints, all tasks are executed to the maximum extent possible without violation of these constraints. To exemplify how tracking of the ante-impact reference defined in Section \ref{sec:reference_generation} translates to such a cost function, consider the ante-impact end effector target acceleration
	\begin{equation}\label{eq:p_t_ante}
	\ddot{\bm p}_{t}^a = \ddot{\bar{\bm p}}^a_{d} + 2 k_p \left(\dot{\bar{\bm p}}^a_{d} - \bm J_p \dot{\bm q}\right) + k_p^2 \left({\bar{\bm p}^a_{d} - \bm p}\right),
	\end{equation}
	with task gain $k_p > 0$.
	When enforcing $\ddot{\bm p} = \ddot{\bm p}^a_{t}$, the tracking error $\bar{\bm p}^a_{d}-\bm p$ will converge to 0 by means of a simple PD control approach. Hence, the ante-impact cost of the position task is formulated as
	\begin{equation}\label{eq:e1}
	\bm e^a_p = \ddot{\bm p} - \ddot{\bm p}^a_{t},
	\end{equation}
	which, combined with \eqref{eq:p_t_ante}, can be rewritten as 
	\begin{equation}\label{eq:pos_task_ante}
	\bm e^a_p = \bm J_{p}\ddot{\bm q} + \bm\eta_p^a,
	\end{equation}
	with
	\begin{equation}\label{eq:eta_p}
	\bm\eta^a_p = \dot{\bm J}_p\dot{\bm q} - \ddot{\bar{\bm p}}^a_{d} - 2 k_p \left(\dot{\bar{\bm p}}^a_{d} - \bm J_p\dot{\bm q}\right) - k_p^2 \left(\bar{\bm p}^a_{d} - {\bm p}\right)
	\end{equation}
	as a known vector independent of optimization variables $\ddot{\bm q}$ and $\bm \tau$, relying solely on constant parameters and measurements of $\bm q$ and $\dot{\bm q}$ at each time step. 
	Formulating a similar cost for the task concerned with tracking the orientation reference $\bar{\theta}^a_{d}$ with task stiffness $k_\theta$, the squared norm of both errors combine to formulate the total cost function as 
	\begin{equation}\label{eq:cost_ante}
	\begin{aligned}
	E_\text{ante} = & w_p\left(\ddot{\bm q}^T \bm J_{p}^T \bm J_{p} \ddot{\bm q} + 2\left.\bm \eta^a_p\right.^T \bm J_{p} \ddot{\bm q}\right) \\ + & w_\theta\left( \ddot{\bm q}^T \bm J_{\theta}^T \bm J_\theta\ddot{\bm q} + 2\eta^a_\theta \bm J_\theta \ddot{\bm q}\right),
	\end{aligned}
	\end{equation}
	with task weights $w_p, w_\theta >0$, which can be selected by the user. Note that the terms independent of the optimization variables have been discarded in \eqref{eq:cost_ante}. 

	The single constraint under which the cost function \eqref{eq:cost_ante} will be minimized, is given by an adapted version of the equations of motion of \eqref{eq:EOM}, removing the contribution of the contact forces $\bm \lambda$. 
	As addressed in \cite{Bouyarmane2019}, different constraints can also be added to ensure that collisions are avoided, and the manipulator complies with the physical limitations of the robot, regarding joint position, velocity and torque limits. For the sake of simplicity, the corresponding constraints are not included here.

	Combining the cost function and constraint, the ante-impact QP is given by
	\begin{equation}
	(\ddot{\bm q}^*, \bm \tau^*) = \min_{\ddot{\bm q}, \bm \tau} E_\text{ante},
	\end{equation}
	s.t.
	\begin{equation}\label{eq:EOM_ante}
	\bm M\ddot{\bm q} + \bm h = \bm S \bm \tau.
	\end{equation}
	The reference torque $\bm \tau^*$ is subsequently sent to the robot at all times $t_k$ separated by time step $\Delta t$, with $\bm q = \bm q(t_k)$, $\dot{\bm q} = \dot{\bm q}(t_k)$.

\subsection{Intermediate mode}\label{sec:int_mode}

	As addressed in Section \ref{sec:introduction}, it is not possible to apply velocity tracking control of either an ante-, or a post-impact reference during the intermediate mode, as on a real robot the exact contact state will not be known and the velocity information is likely unreliable, also due to the presence of unmodeled flexibilities. 
	The challenge in the formulation of this intermediate mode is hence to formulate a QP that enforces full contact to be established without relying on knowledge of the contact state or on joint velocity measurements. Our approach aims to achieve this by applying torque as if the system is still in an ante-impact configuration, with measurements of $\dot{\bm q}$ in the QP replaced by the joint velocities corresponding to the ante-impact reference, given by
	\begin{equation}
		{\dot{\bm q}}_\text{itmd} := \left[\left.{\dot{\bm q}}_{\text{rob},\text{itmd}}\right.^T \ \dot{q}_{4,\text{itmd}}\right]^T,
	\end{equation}
	with
	\begin{equation}
	{\dot{\bm q}}_{\text{rob},\text{itmd}} := \begin{bmatrix}\bm J_{p,\text{rob}} \\ \bm J_{\theta,\text{rob}}\end{bmatrix}^{-1}  \begin{bmatrix} \dot{\bar{\bm p}}^a_{d} \\ \dot{\bar{\theta}}^a_{d}\end{bmatrix},
	\end{equation}
	and $\dot{q}_{4,\text{itmd}}$ given by the nominal estimated ante-impact velocity of the plank. 
	The reasons for making this choice will become apparent when describing the new cost function and constraints below.

	To describe the cost function corresponding to the intermediate mode, we consider first of all the ante-impact end effector position task from \eqref{eq:pos_task_ante}, replacing $\dot{\bm q}$ by $\dot{\bm q}_\text{itmd}$, to give the corresponding error definition
	\begin{equation}\label{eq:pos_task_int}
		{\bm e}^a_{p,\text{itmd}} = \bm J_{p}\ddot{\bm q} + {\bm\eta}_{p}^\text{itmd},
	\end{equation}
	with
	\begin{equation}\label{eq:eta_int}
		{\bm\eta}_{p}^\text{itmd} = {\dot{\bm J}}_{p,\text{itmd}}{\dot{\bm q}_\text{itmd}} - \ddot{\bar{\bm p}}^a_{d} - k_p^2 \left(\bar{\bm p}^a_{d} - {\bm p}\right),
	\end{equation}
	\begin{equation}
	{\dot{\bm J}}_{p,\text{itmd}} := \sum_{i=1}^{4}\frac{\partial \bm J_p}{\partial q_i}{\dot{q}}_{\text{itmd},i}.
	\end{equation}
	Please note that the velocity feedback term $\dot{\bar{\bm p}}^a_{d} - \bm J_p {\dot{\bm q}}_\text{itmd} = \bm 0$ has dropped out in \eqref{eq:eta_int}, leaving only a feedforward term and position feedback term in the cost function. 
	
	Similarly, the cost corresponding to the orientation task can be formulated, leading to the full intermediate mode cost function
	\begin{equation}\label{eq:cost_int}
	\begin{aligned}
	E_\text{itmd} = & w_p\left(\ddot{\bm q}^T \bm J_{p}^T \bm J_{p} \ddot{\bm q} + 2 \left.{\bm\eta}^\text{itmd}_{p}\right.^T \bm J_{p} \ddot{\bm q}\right) \\ + & w_\theta\left( \ddot{\bm q}^T \bm J_{\theta}^T \bm J_\theta\ddot{\bm q} + 2{\eta}^\text{itmd}_{\theta} \bm J_\theta \ddot{\bm q}\right).
	\end{aligned}
	\end{equation}
	Regarding constraints, the equation of motion constraint \eqref{eq:EOM_ante} is modified, again replacing ${\dot{\bm q}}$ by ${\dot{\bm q}}_\text{itmd}$, resulting in the intermediate mode QP formulation 
	\begin{equation}
	(\ddot{\bm q}^*, \bm \tau_\text{act}^*) = \min_{\ddot{\bm q}, \bm \tau_\text{act}}	E_\text{itmd},
	\end{equation}
	s.t.
	\begin{equation}\label{eq:EOM_intermediate}
	\bm M\ddot{\bm q} + \bm h(\bm q,{\dot{\bm q}_\text{itmd}}) = \bm S\bm \tau.
	\end{equation}
	With the formulation of this QP, we are essentially using the same controller as in the ante-impact mode, with an adapted velocity reference as soon as we detect the first impact. This will result in a minimal jump in desired torque when switching from the ante-impact to the intermediate mode, while the position feedback term together with feedforward encourages that full contact will be established during the intermediate mode.

\subsection{Post-impact mode}

	Finally, after the final contact state is established, which is assumed to be identified \cite{Haddadin2017}, the control input is determined through the post-impact QP, in which it is assumed that both contacts remain closed, as common for QP robot control. This can be done by a standard QP \cite{Bouyarmane2019,Salini2010} using the extended post-impact reference $\bar{\bm p}_{d}$ to formulate the cost function related to the position task. While the ante-impact and intermediate mode did not explicitly take the interaction forces $\bm \lambda$ into account, the post-impact QP does so, and hence, $\bm \lambda$ is included in the optimization variables together with $\ddot{\bm q}$ and $\bm \tau$. 

	The cost related to the end effector position tracking task in the post-impact mode can be formulated as was done in the ante-impact mode through \eqref{eq:pos_task_ante}, only swapping the ante-impact reference $\bar{\bm p}^a_{d}$ with the post-impact reference $\bar{\bm p}^p_{d}$, to give
	\begin{equation}\label{eq:pos_task_post}
	\bm e^p_p = \bm J_{p}\ddot{\bm q} + \bm\eta_p^p,
	\end{equation}
	with
	\begin{equation}\label{eq:eta_post_p}
	\bm\eta^p_p = \dot{\bm J}_p\dot{\bm q} - \ddot{\bar{\bm p}}^p_{d} - 2 k_p \left(\dot{\bar{\bm p}}^p_{d} - \bm J_p\dot{\bm q}\right) - k_p^2 \left(\bar{\bm p}^p_{d} - {\bm p}\right).
	\end{equation}
	As addressed in Section \ref{sec:reference_generation}, the end effector orientation in the post-impact mode is not explicitly controlled, since the assumption that $\gamma_1 = \gamma_2 = 0$ implicitly prescribes the end effector orientation for a given end effector position. To ensure a uniquely allocated input resulting from the post-impact QP on top of the 2DOF position task, one can choose to add a regulation task with low weight, adding $\bm \tau^T \bm \tau$ to the cost function. In our case, however, we chose to add a task encouraging an equal contact force distribution over both contact points, with corresponding cost
	\begin{equation}
		e^p_\lambda = \lambda_1 - \lambda_2,
	\end{equation}
	with corresponding weight $w_\lambda$ leading to the total cost function
	\begin{equation}\label{eq:cost_post}
	E_\text{post} = w_p\left(\ddot{\bm q}^T \bm J_{p}^T \bm J_{p} \ddot{\bm q} + 2\left.\bm \eta^p_p\right.^T \bm J_{p} \ddot{\bm q}\right) + w_\lambda (\lambda_1 - \lambda_2)^2.
	\end{equation}

	For the post-impact constraints, first of all, the constraint prescribing the equations of motion should now include the contact forces, hence \eqref{eq:EOM} is included in the QP. Under the assumption that both contacts are closed, with $\bm \gamma = \bm 0$ and $\dot{\bm \gamma} = \bm J_N \dot{\bm q} = \bm 0$, we need to include a constraint that enforces $\ddot{\bm \gamma} = \bm 0$, indicating that both contacts remain closed. 
	Since the contact forces cannot become negative, we also add $\bm \lambda \geq \bm 0$ as a constraint, resulting in the post-impact QP:
	\begin{equation}
	(\ddot{\bm q}^*, \bm \tau^*, \bm \lambda^*) = \min_{\ddot{\bm q}, \bm \tau,\bm\lambda}	E_\text{post},
	\end{equation}
	s.t.
	\begin{align}
		\bm M \ddot{\bm q} + \bm h &= \bm S \bm \tau + \bm J_{N}^T  \bm \lambda, \\
		\bm J_N \ddot{\bm q} + \dot{\bm J}_N \dot{\bm q} &= \bm 0, \\ 
		\bm \lambda &\geq \bm 0.
	\end{align}

\section{Numerical validation}\label{sec:numerical_validation}

	To validate the proposed control approach, simulations have been performed using two different robot models. First, results are presented on simulations that use a rigid robot model with rigid contact model as described in Section \ref{sec:robot_dynamics}. After this, the simulations are repeated for a robot model with flexibility modeled in its joints, and contact described via a compliant contact model. The latter model more closely resembles reality, as flexibility is generally present in, e.g., the drivetrain of robot joints, resulting in oscillations as a result of impacts. These simulations suggest that the developed approach, which uses the assumption of a rigid robot and contact model, is suitable for actual robot control. We will use both simulation models to compare the control approach against two similar tracking control approaches, where we;  1) use the classical adaptation of reference spreading without an intermediate mode as used in \cite{Saccon2014}, where the post-impact reference is tracked as soon as the first impact is detected, and 2) use classical feedback control without reference spreading, and apply the ante-impact mode before the nominal impact time, and the post-impact mode after the nominal impact time. The approach proposed in this paper will be referred to as reference spreading with intermediate mode. In both simulations, the configuration of the plank $q_4$ is initialized with an offset compared to the estimated configuration, which mimics the uncertainty in the environment, and will result in spurious impacts resulting in the system to enter an unexpected contact state. The parameters used in simulations for the rigid and flexible model are given in Table \ref{tab:values}. Large control gains are chosen on purpose to show that we can track the desired motion with high accuracy, while avoiding large peaks in $\bm \tau^*$.
	
\begin{table}[]
	\centering
	\caption{Numerical values used for simulations.}
	\begin{tabular}{l|l}
		Parameter                              & Value                                          \\ \hline
		$(m_1, m_2, m_3)$                      & $(8, 8, 4)$ {[}kg{]}                           \\ \hline
		$(I_{g,1}, I_{g,2}, I_{g,3}, I_{o,4})$ & $(0.03, 0.03, 0.005,4.5)${[}kg m$^2${]}            \\ \hline
		$(L_1, L_2, L_3)$                      & $(0.3, 0.3, 0.15)$ {[}m{]}                     \\ \hline
		$(w_3, w_4, \Delta_x, \Delta_y)$       & $(0.15, 0.04, 0.1, 0.35)$ {[}m{]}              \\ \hline
		$k_4$                                  & $40$ {[}N/m{]}                                 \\ \hline
		$d_4$                                  & $40$ {[}Ns/m{]}                                \\ \hline
		$(k_p, k_\theta)$                      & $(20,20)$ {[}-{]}                              \\ \hline
		$(w_p, w_\theta, w_\lambda)$           & $(1,1,1)$ {[}-{]}                              \\ \hline
		$\bm K$                                & $\text{diag}(30, 30, 15) 10^3$ {[}N/m{]}       \\ \hline
		$\bm D$                                & $\text{diag}(10, 10, 5)$ {[}N/m{]}             \\ \hline
		$\bm B$                                & $\text{diag}(0.24, 0.24, 0.08)$ {[}kg m$^2${]} \\ \hline
		$\bm B_\theta$                         & $0.05 \bm B$                                   \\ \hline
		$c$                                    & $1.5$ {[}-{]}                                  \\ \hline
		$k_\text{env}$                         & $3.2 \ 10^8$ {[}N/m{]}                           \\ \hline
		$d_\text{env}$                         & $3.2 \ 10^{11}$ {[}Ns/m{]}                        
	\end{tabular}
	\label{tab:values}
\end{table}

	\subsection{Numerical results with a rigid robot model}

\begin{figure}
		\centering
		\includegraphics[width=0.88\linewidth]{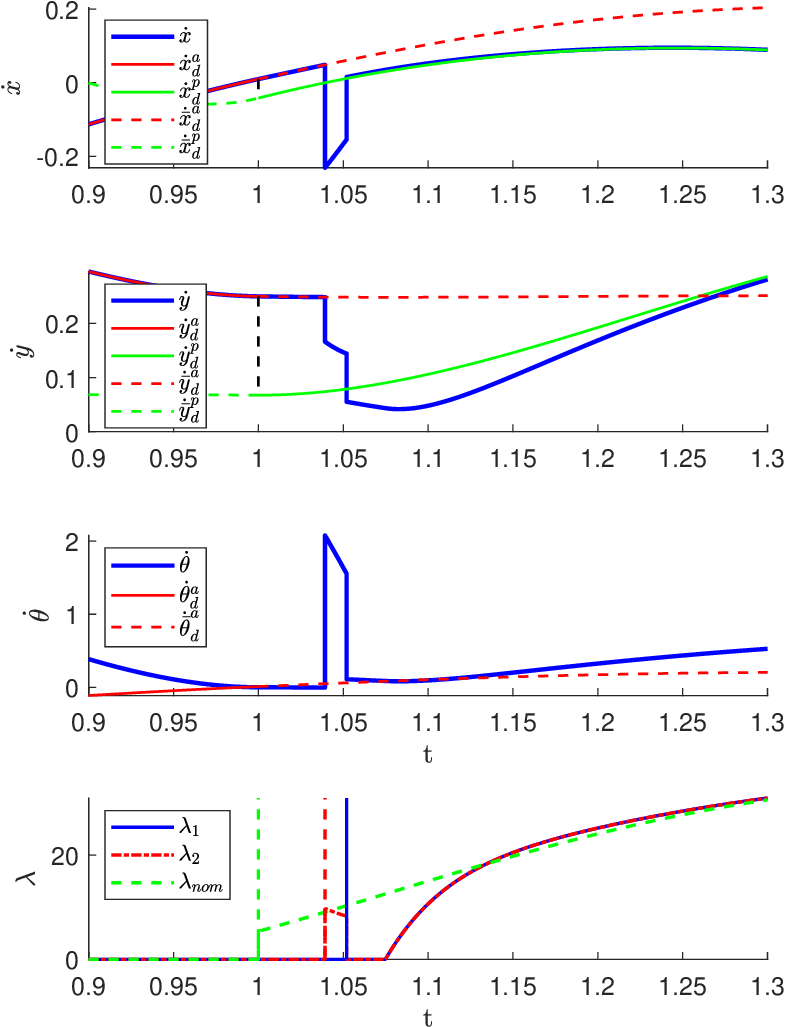}
		\caption{Cartesian linear and angular end effector velocity and contact forces for the rigid robot model, for the reference spreading with intermediate mode control approach proposed in this paper.}
		\label{fig:velocities_rigid}
	\end{figure}

	First, we will show the results of the simulation that employs a rigid robot model with the reference spreading with intermediate mode controller. 
	In Figure \ref{fig:velocities_rigid}, the Cartesian velocities resulting from this simulation are depicted around the nominal impact time, together with the contact forces $\bm \lambda$ against the nominal contact force $\lambda_{\text{nom},1} = \lambda_{\text{nom},2} = \lambda_{\text{nom}}$. In this figure, it can be seen that the extended ante-impact reference is indeed followed until the occurrence of the first impact, indicated by the impulsive contact force $\lambda_2$. As soon as the second impact occurs, full contact is established and the post-impact reference begins to be followed. As expected, we observe two distinct jumps in the velocities corresponding to the impulsive forces. 
	\begin{figure}
		\centering
		\includegraphics[width=0.87\linewidth]{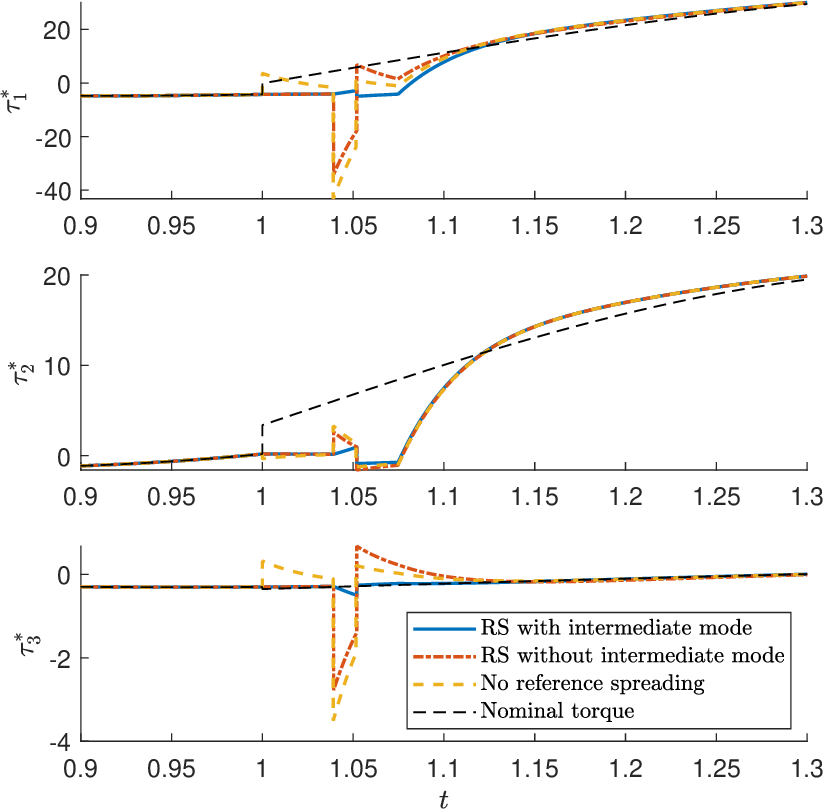}
		\caption{Commanded torque as simulated using a rigid robot model for different control approaches, compared against the nominal torque.}
		\label{fig:torque_rigid}
	\end{figure}
	In Figure \ref{fig:torque_rigid}, the desired torques resulting from the control approach described in Section \ref{sec:control_approach} are shown, both for simulations that use the reference spreading with intermediate mode controller, but also for simulations with a controller that does not use reference spreading, and one that uses reference spreading without intermediate mode. From these simulations, it can be observed that, around the impact time, in both the approach without intermediate mode and without reference spreading, the commanded torques peak as soon as the first impact occurs, and the manipulator enters an undefined contact state. This is because for both these controllers, the post-impact reference is already starting to be tracked, even though full contact has not been established. This results in a temporary large velocity error (that would resolve automatically via the subsequent impact, if left undisturbed), that via the velocity feedback action translates itself into an undesirable large control effort. This undesired behavior is not observed for the reference spreading with intermediate mode controller, as expected, due to the design of the intermediate controller as discussed in Section \ref{sec:int_mode}. 
	
	\subsection{Numerical results with a flexible robot model}

	The flexible robot model used for simulations is a more realistic model describing a robot with flexible joints with corresponding low-level torque control law, presented in \cite{Albu2007}, and is given by
	\begin{align}\label{eq:new_tau}
	\bm \tau_\text{flex} &= \bm K \left(\bm \theta_\text{rob} - \bm q_\text{rob}\right), \\ \label{eq:robot_dyn}
	\bm M \ddot{\bm q} + \bm h &= \bm S \bm \tau_\text{flex} + \bm S \bm D \bm K^{-1} \dot{\bm \tau}_\text{flex} + \bm J_N^T \bm \lambda, \\
	\bm B \ddot{\bm \theta}_\text{rob} &= - \bm \tau_\text{flex} - \bm D \bm K^{-1} \dot{\bm \tau}_\text{flex} + \bm \tau,
	\label{eq:motor_dyn} 
	\end{align}
	with motor inertia $\bm B$, motor-side joint positions $\bm \theta_\text{rob}$, transmission stiffness $\bm K$ and damping $\bm D$, and internal transmission torque $\bm \tau_\text{flex}$. The low level torque control law
	\begin{equation}
		\bm \tau = \bm B \bm B_\theta^{-1} \bm \tau^* + (\bm I- \bm B \bm B_\theta^{-1})(\bm \tau_\text{flex} + \bm D \bm K^{-1}\dot{\bm \tau}_\text{flex}),
	\end{equation}
	with $\bm B_\theta$ as a positive diagonal matrix containing the desired reduced inertia of each joint, is designed to make the robot joint torque $\bm \tau_\text{flex}$ track at best the torque reference $\bm \tau^*$ resulting from one of the three QPs described in Section \ref{sec:control_approach}.  
	The contact force $\bm \lambda$ appearing in \eqref{eq:robot_dyn} is determined using a compliant contact model, based on the exponentially extended Hunt-Crossley model \cite{Carvalho2019}, and is given by
	\begin{equation}
	\lambda_i = \left\{
	\begin{aligned}
	\mathcal{K}(\dot{\gamma}_i)(-\gamma_i)^c \quad & \text{if} \ \gamma_i \leq 0,  \\
	0 \quad & \text{if} \ \gamma_i > 0, 
	\end{aligned}
	\right.
	\end{equation}
	where negative $\gamma_i$ denotes compenetration of the contacting surfaces, $c$ is a geometry dependent parameter, and
	\begin{equation}
	\mathcal{K}(\dot{\gamma}_i) = \left\{
	\begin{aligned}
	k_\text{env} - d_\text{env}\dot{\gamma}_i \quad & \text{if} \ \dot{\gamma}_i \leq 0,  \\
	k_\text{env} \exp{\left(-\frac{d_\text{env}}{k_\text{env}}\dot{\gamma}_i\right)} \quad & \text{if} \ \dot{\gamma}_i > 0, 
	\end{aligned}
	\right.
	\end{equation}
	denotes the velocity-dependent contact stiffness, with stiffness and damping parameters $k_\text{env}$ and  $d_\text{env}$ respectively.

	For this flexible model, switching from the ante-impact to the intermediate mode occurs as soon as the first contact is closed (i.e. $\exists i \ (\gamma_i \leq 0)$), while the post-impact mode is entered as soon as sustained contact is made at both contact points (i.e. $\forall i \ (\gamma_i \leq 0 \land \left|\dot{\gamma}_i\right| \leq \epsilon)$) with threshold $\epsilon$. In real-life robotic applications, switching will occur based on impact detection, for example through jump-aware filtering \cite{Rijnen2018a}, as is used in the experimental validation of reference spreading in \cite{Rijnen2020}.

\begin{figure}
		\centering
		\includegraphics[width=0.915\linewidth]{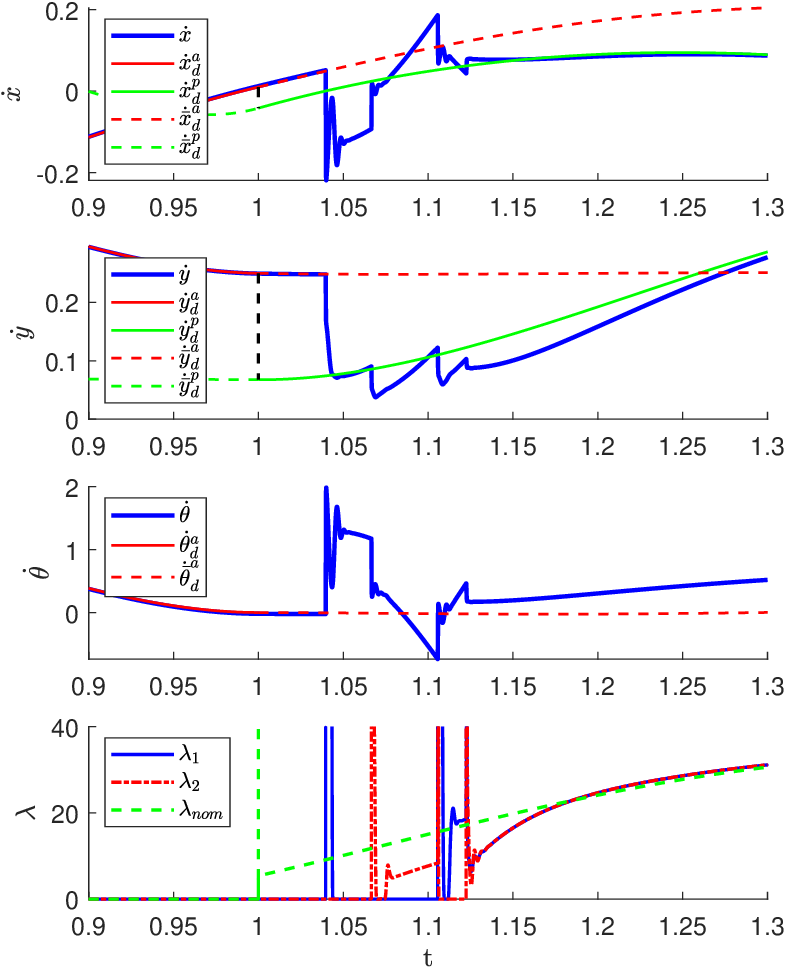}
		\caption{Cartesian linear and angular end effector velocity and contact forces for the flexible robot model, for the reference spreading with intermediate mode control approach.}
		\label{fig:velocities_flex}
	\end{figure}

	Using the flexible model, simulations have been performed with the same control strategies as for the rigid robot model. For the reference spreading with intermediate mode control approach, the Cartesian velocities and contact forces are depicted in Figure \ref{fig:velocities_flex}. The effects of including joint flexibility and compliant contacts in this model can clearly be observed, as large vibrations in the velocity occur after the first impact, combined with several impulsive contact force peaks. This reinforces why, during the intermediate mode, which is active until full and sustained contact is established, we chose not to rely on velocity measurements of the system. As a result of the intermediate mode, sustained contact is in the end obtained, after which the post-impact mode is entered, where the control performance remains almost identical to that observed in Figure \ref{fig:velocities_rigid}.
	
	\begin{figure}
		\centering
		\includegraphics[width=0.915\linewidth]{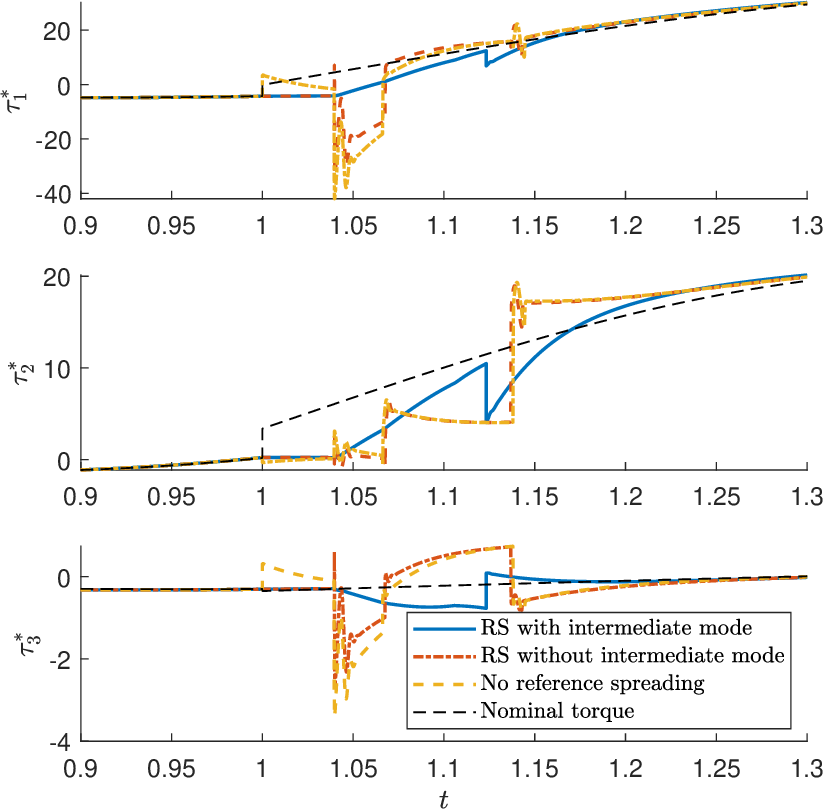}
		\caption{Commanded torque as simulated using a flexible robot model with different control approaches, compared against the nominal torque.}
		\label{fig:torque_flex}
	\end{figure}

	Comparing the commanded torques for the three aforementioned control approaches against the nominal torque, depicted in Figure \ref{fig:torque_flex}, the beneficial effects of not utilizing any velocity information in the intermediate mode of the proposed control approach become apparent, since the vibrations resulting from the impact do not translate to $\bm \tau^*$, as opposed to the other two control approaches. Application of the reference spreading with intermediate mode controller hence leads to a predictable and reliable reference torque used to complete the contact, before continuing to the post-impact QP where the corresponding reference is again followed.

\section{Conclusion}\label{sec:conclusion}

	In this paper, an improved formulation of reference spreading control for tracking trajectories with nominally simultaneous impacts has been presented. Compared to previous formulations, we show that reference spreading can be cast within the framework of QP control, and that position feedback can be employed in the intermediate mode. 
	Based on simulation results on a realistic robot model with flexible joints, the approach can be used to track a desired reference before and after a nominally simultaneous impact occurs. An intermediate control mode is defined to establish full contact when contact is only partially established, without relying on information of the contact state or on velocity measurements. The controller in the intermediate mode is of the same structure as that of the ante-impact control mode, resulting in a reliable torque signal that can be used to complete the contact. Future work involves scaling up to a more realistic 3D case and performing a corresponding experimental study on real-life robotic systems.

\addtolength{\textheight}{-0cm}   



\section*{APPENDIX}


	\subsection{Formulation nominal trajectory coherent with impact map} \label{app:appendix_reference}

	Here, we explain how to formulate a post-impact reference ${\bm p}^p_{d}(t)$ that is consistent with the ante-impact reference ${\bm p}^a_{d}(t)$ and ${\theta}^a_{d}(t)$. As $\bm q^- = \bm q^+$, it is straightforward that ${\bm p}^a_{d}(t_\text{imp}) = {\bm p}^p_{d}(t_\text{imp})$. 
	To formulate a reference that also adheres to the impact map on a velocity level, a requirement is to determine a unique impact configuration $\bm q^-$ corresponding to the operational space reference $({\bm p}^a_{d}(t_\text{imp}), {\theta}^a_{d}(t_\text{imp}))$. Assuming that the ante-impact state of the plank, given by $(q_4^-, \dot{q}_4^-)$, is known, there are still 2 solutions left to the inverse kinematics problem, either an elbow up or an elbow down configuration. However, tracking this reference from a given initial condition will result in a unique impact configuration $\bm q^-$, assuming the robot stays away from singularities.

	With the nominal $\bm q^- = \bm q^+$ determined through the aforementioned process, inverse velocity kinematics can be used to formulate the nominal corresponding $\dot{\bm q}^-$, as
	\begin{equation}
	\dot{\bm q}^- = \left[\left.\dot{\bm q}_\text{rob}^-\right.^T \ \dot{q}_4^-\right]^T,
	\end{equation}
	with
	\begin{equation}
	\dot{\bm q}_\text{rob}^- = \begin{bmatrix}\bm J_{p,\text{rob}}(\bm q^-) \\ \bm J_{\theta,\text{rob}}(\bm q^-)\end{bmatrix}^{-1}  \begin{bmatrix} \dot{{\bm p}}^a_{d}(t_\text{imp}) \\ \dot{{\theta}}^a_{d}(t_\text{imp})\end{bmatrix}.
	\end{equation}
	and $\dot{q}_4^-$ assumed to be known.  
	Through \eqref{eq:impact_map}, a value for $\dot{\bm q}^+$ can then be determined, which leads to
	\begin{equation}
	{\bm p}^p_{d}(t_\text{imp}) = \bm J_p(\bm q^+)\dot{\bm q}^+
	\end{equation}
	as a requirement for the post-impact reference to be consistent with the impact map given the ante-impact reference.

\section*{ACKNOWLEDGMENT}

This work was partially supported by the Research Project I.AM. through the European Union H2020 program under GA 871899.


\bibliography{References/library}{}
	\bibliographystyle{ieeetr}

\end{document}